\documentclass{article}

\usepackage[preprint]{neurips_2024}

\usepackage[utf8]{inputenc} 
\usepackage[T1]{fontenc}    
\usepackage{hyperref}       
\usepackage{url}            
\usepackage{booktabs}       
\usepackage{amsfonts}       
\usepackage{nicefrac}       
\usepackage{microtype}      
\usepackage{xcolor}         

\usepackage{amsmath}
\usepackage{graphicx}
\usepackage{tabularx}
\usepackage{bm}
\usepackage{multirow}
\usepackage{subfigure}

\title{Pre-Trained Vision-Language Models as Partial Annotators}

\author{%
  Qian-Wei Wang$^{1,2}$, Yuqiu Xie$^1$, Letian Zhang$^1$, Zimo Liu$^2$, Shu-Tao Xia$^{1,2}$\thanks{Corresponding author.} \\
  $^1$Tsinghua Shenzhen International Graduate School, Tsinghua University \\
  $^2$Research Center of Artificial Intelligence, Peng Cheng Laboratory\\
  \texttt{\{wanggw21, zbm18, zmy20, litx21\}@mails.tsinghua.edu.cn}\\ \texttt{liuzm@pcl.ac.cn, xiast@sz.tsinghua.edu.cn} \\
}

\begin{document}

\maketitle

\begin{abstract}
Pre-trained vision-language models learn massive data to model unified representations of images and natural languages, which can be widely applied to downstream machine learning tasks. In addition to zero-shot inference, in order to better adapt pre-trained models to the requirements of downstream tasks, people usually use methods such as few-shot or parameter-efficient fine-tuning and knowledge distillation. However, annotating samples is laborious, while a large number of unlabeled samples can be easily obtained. In this paper, we investigate a novel "pre-trained annotating - weakly-supervised learning" paradigm for pre-trained model application and experiment on image classification tasks. Specifically, based on CLIP, we annotate image samples with multiple prompt templates to obtain multiple candidate labels to form the noisy partial label dataset, and design a collaborative consistency regularization algorithm to solve this problem. Our method simultaneously trains two neural networks, which collaboratively purify training labels for each other and obtain pseudo-labels for self-training, while adopting prototypical similarity alignment and noisy supervised contrastive learning to optimize model representation. In experiments, our method achieves performances far beyond zero-shot inference without introducing additional label information, and outperforms other weakly supervised learning and few-shot fine-tuning methods, and obtains smaller deployed models. Our code is available at: \url{https://anonymous.4open.science/r/Co-Reg-8CF9}.
\end{abstract}

\section{Introduction}

With the invention of effective multi-modal deep frameworks and training algorithms, as well as significant improvements in computing power, pre-trained vision-language models \cite{li2022blip, bao2022vlmo, achiam2023gpt} have demonstrated impressive capabilities on a wide range of tasks. Taking CLIP (Contrastive Language-Image Pre-Training) \cite{radford2021learning} as an example, it models unified representations of images and natural language by learning massive "image-text" pairs. During pre-training, semantically related pairs of images and texts are encoded by the image encoder and the text encoder to obtain aligned representations. Because CLIP learns from natural language descriptions, it can leverage the rich contextual information provided by language to understand and categorize new images without having been explicitly trained on them. When presented with a new downstream task, CLIP can interpret the category descriptions which are then matched by the encoded input images in the shared embedding space. 

There are two main issues when directly applying CLIP to downstream image classification tasks: 1. For tasks where the input image domain differs significantly from the general image domain, CLIP often fails to accurately match the input images with the corresponding category descriptions, resulting in poor performances; 2. In scenarios with limited computational resources, the overhead of executing large model inference may be unsustainable. Addressing on these problems, previous researchers usually perform fine-tuning on downstream datasets \cite{wortsman2022robust} or distill pre-trained knowledge \cite{gou2021knowledge} to the deployed model. However, both of them require additional human annotations, which are costly to obtain. At the same time, unlabeled samples can be obtained at extremely low cost, motivating us to study methods for applying pre-trained models without the need for additional human annotation.

In this paper, we investigate a novel paradigm for the deployment of pre-trained vision-language models, called "pre-trained annotating - weakly-supervised learning (P-WSL)", in which we utilize the pre-trained models as weak annotators to annotate unlabeled samples for downstream tasks and subsequently employ weakly-supervised learning \cite{zhou2018brief} algorithms to train on these annotations. Table \ref{paradigms} compares P-WSL with other mainstream paradigms. It is evident that P-WSL is the only one that can achieve performance improvements over the original model without using manual annotations. Meanwhile, by retraining dedicated small models on the downstream samples, the inference model size are significantly reduced. Additionally, due to the fact that few-shot fine-tuning techniques (e.g., prompt learning \cite{zhou2022learning} and adaptors \cite{chen2023vision, gao2024clip}) only add a small number of trainable parameters to the pre-trained models, their performance improvements are usually limited.

Since each prompt template corresponds to a classifier, in practice, we consider the predicted category of each prompt template as a "candidate". We combine all candidate labels for a sample into a collection and model it as a noisy partial label learning problem (NPLL), and design a \textbf{Co}llaborative consistency \textbf{Reg}ularization (Co-Reg) method to solve it. Specifically, our method simultaneously trains two neural networks, which collaboratively purify training labels for each other and obtain pseudo-labels for self-training, while adopting prototypical similarity alignment and noisy supervised-contrastive learning to optimize model representation. 

Our main contributions can be summarized as:
\begin{itemize}
    \item One of the pioneering works applying weakly-supervised methods to large model scenarios, introducing the P-WSL paradigm for applying pre-trained models. It is worth noting that this paradigm is not only applicable to vision-language models but can also be extended to other types of pre-trained models, such as large language models.

    \item Propose a collaborative consistency regularization method, in which training labels are collaborative purified and the model's representation ability is further enhanced by mining representation-level consistency. Extensive experiments demonstrate its effectiveness.
    
    \item A unified comparison method for different types of weakly-supervised problems is provided. Previously, methods dealing with different kinds of weak supervision were typically compared within synthetic datasets following their own settings. By leveraging vision-language models, datasets can be annotated according to different types of weak supervision, followed by executing corresponding algorithms.
    
\end{itemize}

\begin{table}[]
\centering
\resizebox{0.95\textwidth}{!}{
\begin{tabular}{@{}ccccc@{}}
\toprule
Paradigms               & Samples      & Human Annotations & Inference Size   & Perf. Improvements \\ \midrule
Zero-Shot               & $\times$     & $\times$          & -                & -                  \\
Prompt Learning / Adapter & few          & few               & increase sightly & $\checkmark$       \\
KD$_{unsupervised}$       & $\checkmark$ & $\times$          & small            & $\times$           \\
KD$_{supervised}$         & $\checkmark$ & $\checkmark$      & small            & $\checkmark$       \\
Fully Fine-Tuning       & $\checkmark$ & $\checkmark$      & -                & $\checkmark$       \\
P-WSL                   & $\checkmark$ & $\times$          & small            & $\checkmark$       \\ \bottomrule
\end{tabular}}
\caption{Comparison among different pre-trained model application paradigms. KD$_{supervised}$ and KD$_{unsupervised}$ represent knowledge distillation with or without task labels, respectively. "-" means remaining the same with original model.}
\label{paradigms}
\end{table}

\section{CLIP Annotated Partial Labels}\label{annotation}
This section introduces how we use pre-trained vision-language model to annotate downstream image datasets and obtain training labels. We take CLIP \cite{radford2021learning} as the example, and our experiments are also based on it. 

We use a collection of prompt templates, denoted as $\{\mathcal{T}_1(\cdot), \mathcal{T}_2(\cdot) \dots \mathcal{T}_d(\cdot)\}$, and combine them with the class names of images to form the textual input. The template here is like: "a photo of a \{\}.", where \{\} is replaced with class names. We denote the class names as $\{n_1, n_2, \dots, n_C\}$, and denote the combination of $j$-th class and $i$-th template as $\mathcal{T}_i(n_j)$.

For each template $\mathcal{T}_i(\cdot)$, we combine it with all class names and then take them as the input of CLIP text encoder and obtain $C$ textual representations $\{t_1, t_2, \dots, t_C\}$, where $t_j = \textrm{TextEncoder}(\mathcal{T}_i(n_j))$. Input the training image into the CLIP Image Encoder and obtain image representation $r$. Then, we can predict the probabilities of the image belonging to different classes under this prompt template as $p_i = \textrm{softmax}(r \cdot t_1, r \cdot t_2, \dots, r \cdot t_C)$. 

At this time, we face two possible solutions. One is calculating the average of predicted probabilities, and select the class with largest predicted probability as the annotated single label, which is the regular noisy label learning scenario. The second is obtaining the one-hot label of each predicted probabilities $\hat{p}_i = \textrm{onehot}(p_i)$, and deeming each one-hot label as a candidate and forming the partial label $Y = (y_1, y_2, \dots, y_C) \in \{0, 1\}^C$, in which $y_j = \mathbb{I}(j \in S)$ and $S = \textrm{set}(\hat{p}_i), 1 \leq i \leq d$ denotes the candidate label set. Here, $\mathbb{I}(\cdot)$ returns $1$ if the inside condition is met or otherwise $0$ and $\textrm{set}(\cdot)$ forms the inputs into a set.

We found that the second solution achieves better results especially under extreme circumstances when most prompt templates fail to provide satisfactory predictions. And at this time, as long as one prompt template makes a correct prediction, the prediction will be included in the candidate label set and the difficult of the algorithm to recognize it as the correct label with the help of consistency regularization is greatly decreased. This is very helpful when the characteristic of downstream task is unknown and prompt engineering can hardly be performed. In this article, we study the pipeline of using partial labels annotated by pre-trained models, and derive an effective NPLL algorithm.

\begin{figure}
    \centering
    \includegraphics[width=1\linewidth]{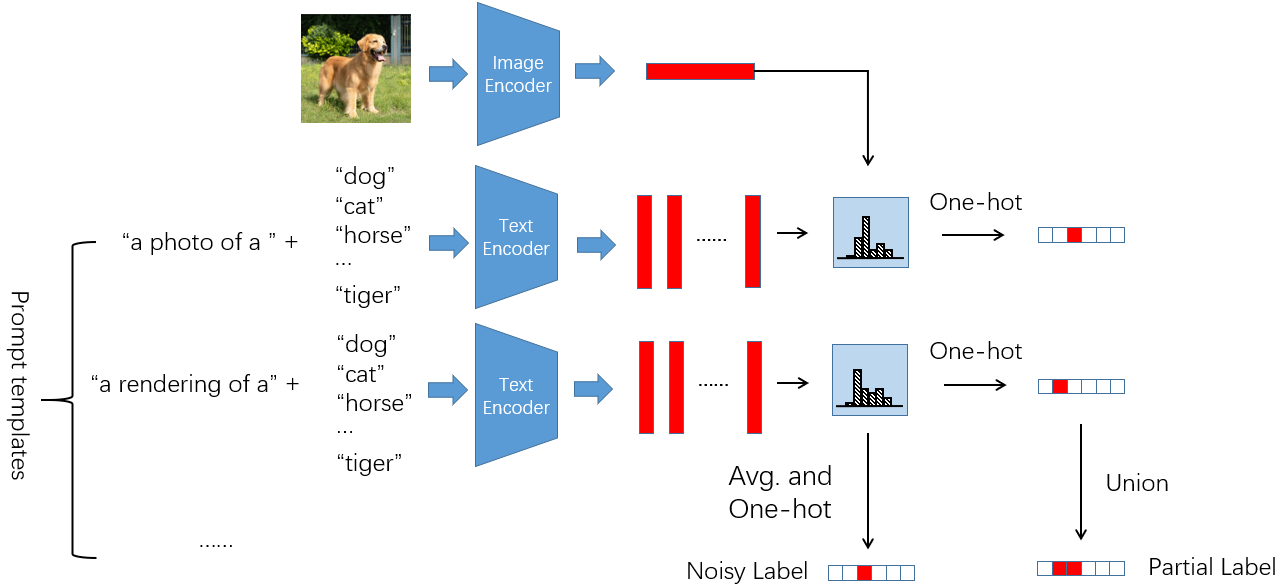}
    \caption{Illustration of how pre-trained annotated single/partial labels are generated.}
    \label{fig:partial}
\end{figure}

\section{Methodology}

\subsection{Co-Pseudo-Labeling}
Our method simultaneously trains two neural networks Net1 and Net2 (denoted as $f(x; \theta_1)$ and $f(x; \theta_2)$), which collaboratively purify training labels for each other and obtain pseudo-labels for self-training and representation refinement. In the next, we take the training of Net2 as an example (the same applies to Net1).

Based on previous successful experiences\cite{DBLP:conf/nips/SohnBCZZRCKL20, berthelot2019remixmatch}, our method adopts two types of data augmentation, i.e., weak data augmentation $\textrm{Aug}_w(\cdot)$ and strong data augmentation $\textrm{Aug}_s(\cdot)$. We take images applied with weak data augmentation (also called weakly-augmented) as input to obtain pseudo-labels, which are used to train images applied with strong data augmentation.

We firstly warm up both networks for few epochs. We adopt the partial cross-entropy loss as the supervised loss for warming up, as well as the negative entropy to prevent from overfitting.
\begin{align}
    L_{sup} = - \log{\sum_{j=1}^{C} y_j f_j(x)}&, \quad L_{reg} = \sum_{j=1}^{C} f_j(x)\log{f_j(x)}, \\
    L_{warm} &= L_{sup} + L_{reg}. 
\end{align}
After warming up, we attempt to classify the provided partial labels as valid or not, i.e., whether the ground-truth labels are in the candidate label sets, based on the minimal-loss criterion \cite{arazo2019unsupervised, chen2019understanding}. Drawing inspiration from DivideMix \cite{li2020dividemix}, we utilize the partitioning from Net1 to train Net2. The minimal-loss criterion assumes that noise-free samples are easier to learn. Under partial label circumstance, we speculate that if the partial label of the sample is valid, the model warm-up trained using supervised loss can predict the sample to a category within its candidate label set with a greater probability. Specifically, we calculate the following loss over the predicted probabilities of all weakly-augmented samples $\{L_{div}(\textrm{Aug}_w(x^i); \theta_1)\}_{i=1}^{N}$.
\begin{equation}
    L_{div}(x; \theta) = - \log{f_j(x; \theta)}, \quad j = \underset{j \in \mathcal{Y}, y_j = 1}{\arg\max} \ f_j(x; \theta_1),
\end{equation}
where $f_j(x; \theta)$ indicates the predicted probability on the $j$-th category of neural network with parameter $\theta$, $\mathcal{Y}$ represents the label space and $N$ is total number of training samples.

We use a two-component Gaussian mixture model (GMM)\cite{permuter2006study} to fit the above losses, and divide the whole training set into a partial split $\mathcal{P} = \{(x^i, p^i) | i \in (1, 2, \dots, N_p)\}$, whose partial labels are assumed to be valid with a probability $w^i$, and an unlabeled split $\mathcal{U} = \{(x^i) | i \in (N_p+1, N_p+2, \dots, N_p+N_u)\}$, whose partial labels are assumed to be non-valid and discarded, according to the prediction from GMM. Here, $p^i = (p^i_1, p^i_2, \dots, p^i_C)$ is the predicted label distribution of $x^i$ from Net1 after re-scaling with Eq.\ref{rescale},
\begin{equation}
    p^i_j = \frac{y^i_j \cdot f_j(\textrm{Aug}_w(x^i); \theta_1)}{\sum y^i f(\textrm{Aug}_w(x^i); \theta_1)}, \quad \text{for } j = 1, 2, \dots, C.
    \label{rescale}
\end{equation}
Then, we combine the predicted label distributions from both networks to obtain the fused pseudo-labels $p'$ with Eq.\ref{label_fuse}, which are then temperature sharpened with Eq.\ref{sharpen} for the subsequent self-training and representation refinement of Net2.
\begin{equation}
p^{'i} = 
\begin{cases}
    w^i \cdot p^i + (1 - w^i) \cdot \bar{p}^i_2, & \text{if } i \in (1, 2, \dots, N_p); \\
    (\bar{p}^i_1 + \bar{p}^i_2) / 2 & \text{if } i \in (N_p+1, N_p+2, \dots, N_p+N_u).
\end{cases}
\label{label_fuse}
\end{equation}
\begin{equation}
    \tilde{p}^i = \frac{(p^{'i}_j)^{1/T}}{\sum (p^{'i})^{1/T}}, \quad \text{for } j = 1, 2, \dots, C.
    \label{sharpen}
\end{equation}
Here, $\bar{p}^i_1 = \frac{1}{K} \sum_{k=1}^K f(\textrm{Aug}_w(x^i); \theta_1)$, $\bar{p}^i_2 = \frac{1}{K} \sum_{k=1}^K f(\textrm{Aug}_w(x^i); \theta_1)$ represent the average predicted probabilities of $K$ weakly-augmented inputs of Net1 and Net2, respectively.

\subsection{Self-Training}
After obtaining the pseudo-labels, we perform self-training on the samples in both the partial split $\mathcal{P}$ and the unlabeled split $\mathcal{U}$. We choose the cross-entropy loss as the training objective on $\mathcal{P}$ and the mean square error on $\mathcal{U}$ due to its noisy-tolerant property:
\begin{equation}
L_{\textrm{self}}(x; \theta) = 
\begin{cases}
    - \sum_{j=1}^C \tilde{p}_j \log{f_j(x; \theta)}, \quad & \text{if } x \in \mathcal{P}; \\
    ||\tilde{p}-f(x; \theta)||_2^2, \quad & \text{if } x \in \mathcal{U}.
\end{cases}
\end{equation}

\subsection{Representation Refinement}
\begin{figure}
    \centering
    \includegraphics[width=1.0\linewidth]{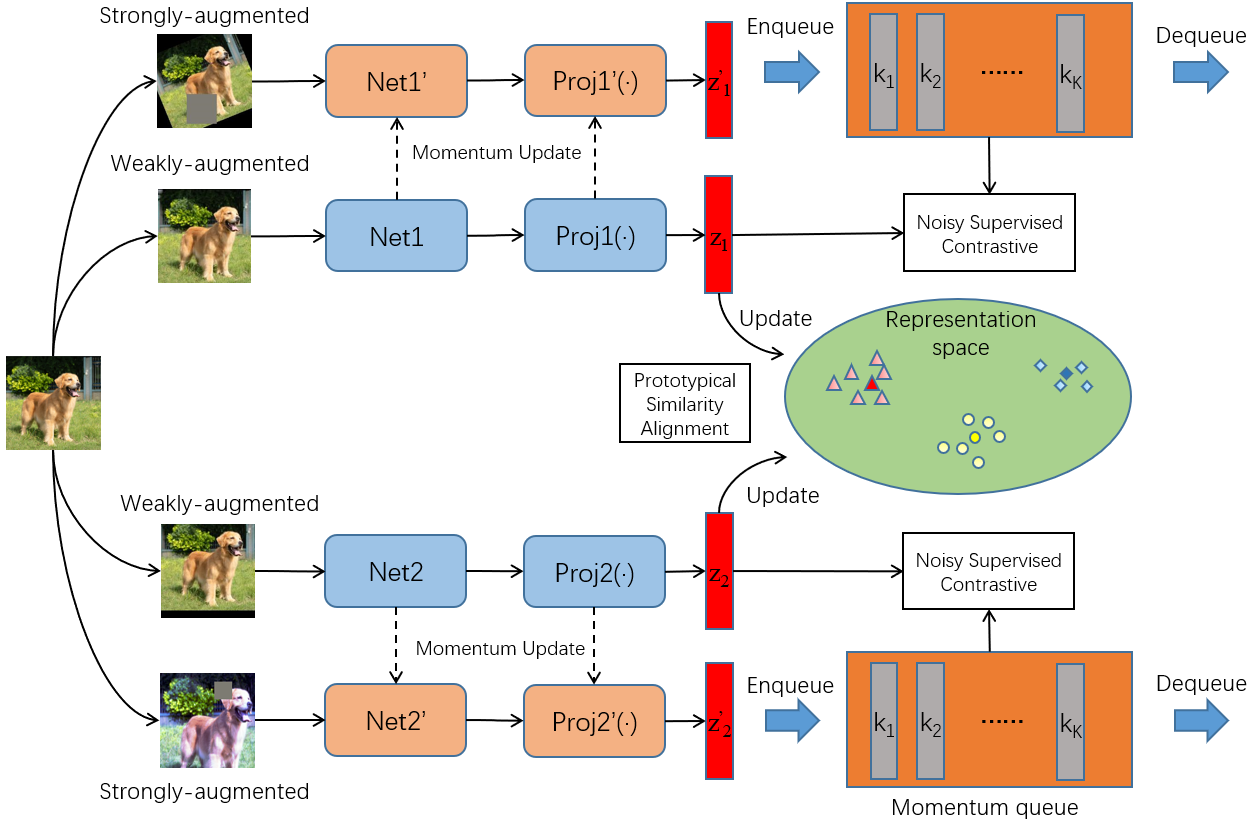}
    \caption{Prototypical similarity alignment and noisy supervised contrastive learning of Co-Reg.}
    \label{fig:cont}
\end{figure}

To further exploit from the data distribution property of downstream unlabeled images while enhancing the model's representation ability, we employ the prototypical similarity alignment and noisy supervised contrastive learning. 

Prototypical similarity alignment enforces consistency between the representation distribution and the label distribution. We project the output representations of image $x^i$ of the two neural networks to a shared embedding space through a two-layer MLP with L2 normalization, respectively (See Fig.\ref{fig:cont}), obtaining $z^i = g(f(\textrm{Aug}_w(x^i); \bar{\theta}))$, in which $g(\cdot)$ represents the MLP projector and $\bar{\theta}$ represents the neural network parameters $\theta$, excluding the last fully-connected layer. During the training process, we maintain a cluster center for each category in the shared representation space that represents the representation of the category, called "prototype", denoted by $\{o_j\}_{j=1}^C$. We believe that different data augmentation variants of the same sample should maintain consistent distributions between label space and representation space. Specifically, our method calculate the similarity distribution over the representation of current image and class prototypes as $s^i = \textrm{softmax}(z^i \cdot o_1, z^i \cdot o_2, \dots, z^i \cdot o_C)$, which is then aligned to the pseudo-label distribution $\tilde{p^i}$. We choose KL-Divergence for samples in partial split $\mathcal{P}$, which have a much higher pseudo-accuracy and mean square error for samples in unlabeled split $\mathcal{U}$. The loss functions for prototypical similarity alignment can be written as:
\begin{equation}
L_{\textrm{prot}}(x; \theta) = 
\begin{cases}
    \sum_{j=1}^C \tilde{p}_j^{T'} \ \log (\tilde{p}_j^{T'} / s_j^{T'}), \quad & \text{if } x \in \mathcal{P}; \\
    ||\tilde{p} - s||_2^2, \quad & \text{if } x \in \mathcal{U}.
\end{cases}
\end{equation}
The class prototypes are momentum updated during training with Eq.\ref{update}.
\begin{equation}
    o_j = \gamma o_j + (1 - \gamma) z^i, \quad j = \underset{j \in \mathcal{Y}}{\arg\max} \ \tilde{p}_j
    \label{update}
\end{equation}
In addition, we utilize contrastive learning to pull together representations of samples from the same class while pushing apart those from different classes, enabling the model to encode more discriminative features on downstream data. In implementation, we adopt the MoCo \cite{DBLP:conf/cvpr/He0WXG20} framework, in which a large-size "first-in-first-out" queue of image representations encoded by the momentum updated copy of our model is maintained. We select positive and negative examples for the current image representation from the representation queue by their pseudo-labels and optimize the following noisy-tolerant contrastive loss:
\begin{equation}
\begin{aligned}
\mathcal{L}_{\textrm{ncont}}(x; \theta) = - \frac{1}{|P(x)|} \sum_{z_+ \in P(x)} \log \frac{\exp(z^\top z_+ / T'')}{\sum\limits_{z_+ \in P(x)} \exp(z^\top z_+ / T'') + \sum\limits_{z_- \in N(x)} \exp(z^\top z_- / T'')},
\end{aligned}
\end{equation}
where $P(x)$ and $N(x)$ separately denote the set of selected positive and negative examples for image $x$, $T'' \geq 0$ is the temperature. We treat $x \in \mathcal{P}$ as confident samples and $x \in \mathcal{U}$ as lacking of confidence and applying the selection strategy for $P(x)$ and $N(x)$ in \cite{wang2024controller}.

Finally, the overall training objective is: 
\begin{equation}
    L = L_{\textrm{self}} + \lambda_1 L_{\textrm{prot}} + \lambda_2 L_{\textrm{ncont}}
\end{equation}

\section{Experiments}
\subsection{Experimental Setup}
\begin{table}[]
\centering
\resizebox{0.85\textwidth}{!}{
\begin{tabular}{@{}cccccccc@{}}
\toprule
Datasets              & CIFAR-10 & CIFAR-100 & SVHN    & FMNIST  & EuroSAT & FER2013 & GTSRB   \\ \midrule
Avg. Candi & 1.39     & 2.36      & 2.41    & 1.58    & 3.26    & 2.38    & 2.84    \\
$\eta$   & 4.79\%   & 21.50\%   & 61.61\% & 22.35\% & 32.89\% & 21.49\% & 58.22\% \\ \bottomrule
\end{tabular}}
\label{stat}
\end{table}

We conduct experiments on several image classification benchmarks: CIFAR-10, CIFAR-100 \cite{krizhevsky2009learning}, SVHN \cite{goodfellow2013multi}, Fashion-MNIST \cite{xiao2017fashion}, EuroSAT \cite{helber2019eurosat}, FER2013 \cite{goodfellow2013challenges} and GTSRB \cite{houben2013detection}. For CIFAR-10, CIFAR-100, Fashion-MNIST and EuroSAT, we use the class descriptions from PyTorch; and for SVHN, FER2013 and GTSRB, their class descriptions (see Section \ref{names}) are manually assigned and are the same for all comparison methods. The prompt templates we used are listed in \ref{prompts}.

We compare the performances of four model application paradigms: zero-shot, unsupervised knowledge distillation (KD), P-WSL and few-shot fine-tuning. The former three requires no human labeling while few-shot fine-tuning is performed with 2, 4, 8, and 16 labeled samples per class. 

We compare our method with four methods under P-WSL paradigms, CR-DPLL \cite{wu2022revisiting}, DivideMix \cite{li2020dividemix}, ALIM-Onehot and ALIM-Scale \cite{xu2024alim}, in which CR-DPLL is for partial label learning, DivideMix is for learning for noisy labels, ALIM-Onehot and ALIM-Scale are for NPLL. As illustrated in Section \ref{annotation}, we first annotate the training images of the above datasets with noisy single/partial labels with CLIP and then perform corresponding methods. And we choose prompt learning method CoOp \cite{zhou2022learning} as the few-shot fine-tuning compared method. For zero-shot and DivideMix, we present the performances with single prompt template "a photo of a \{\}" and using the average of predicted probabilities of multiple prompt templates (superscript with asterisk).

The average amount of candidate labels per training sample and the probabilities of ground-truth label being outside of the candidate sets, i.e. $\eta$, is recorded for partially annotated datasets in Table \ref{stat}.

The pre-trained vision-language model for our experiments is: CLIP ViT-B/32. We use the PreAct ResNet-18 \cite{he2016identity} as the backbone for our method. The training batch-size is 256, and the number of warm up and total epochs are chosen from 50 or 100 and 100 or 800, respectively. The number of weakly-augmented inputs for co-pseudo-labeling is $K = 2$ and the sharpening temperature is $T = 0.5$. The dimension of projected representations is 128, and the length of MoCo queue is 8192. The loss weights are $\lambda_1 = 0.1$, $\lambda_2 = 0.1$ The experiments are all carried on NVIDIA 3090 GPUs.

\subsection{Main Results}

\begin{table}[]
\centering
\setlength{\tabcolsep}{4pt}
\resizebox{0.9\textwidth}{!}{
\begin{tabular}{@{}cccccccc@{}}
\toprule
Methods            & CIFAR-10 & CIFAR-100 & SVHN    & FMNIST  & EuroSAT & FER2013 & GTSRB   \\ \midrule
Zero-Shot$_{train}$  & 88.40\%  & 61.83\%   & 9.34\%  & 62.57\% & 32.26\% & 41.47\% & 24.87\% \\
Zero-Shot$^*_{train}$ & 89.09\%  & 62.75\%   & 9.33\%  & 65.81\% & 30.61\% & 46.55\% & 25.53\% \\
Zero-Shot$_{test}$   & 88.51\%  & 61.55\%   & 8.63\%  & 61.46\% & 31.49\% & 42.07\% & 25.14\% \\
Zero-Shot$^*_{test}$  & 89.01\%  & 62.74\%   & 8.82\%  & 65.14\% & 30.78\% & 46.72\% & 25.57\% \\ \midrule
KD                 & 87.74\%  & 56.80\%   & 8.21\%  & 67.60\% & 34.13\% & 46.43\% & 26.98\% \\ \midrule
CR-DPLL            & 84.20\%  & 60.05\%   & 6.82\%  & 71.27\% & 8.85\%  & 16.75\% & 27.16\% \\
DivideMix          & 93.32\%  & 65.76\%   & 16.46\% & 71.60\% & 42.41\% & 44.31\% & 30.07\% \\
DivideMix$^*$         & 93.83\%  & 66.03\%   & 17.16\% & 74.21\% & 37.74\% & 47.42\% & 32.69\% \\
ALIM-Onehot        & 93.18\%  & 64.60\%   & 17.06\% & 72.42\% & 34.57\% & \textbf{52.77}\% & 31.06\% \\
ALIM-Scale         & 93.59\%  & 64.61\%   & 20.66\% & 72.36\% & 37.11\% & 52.51\% & 31.81\% \\
Co-Reg             & \textbf{94.06}\%  & \textbf{71.04}\%   & \textbf{46.57}\% & \textbf{76.28}\% & \textbf{65.54}\% & 50.26\% & \textbf{41.18}\% \\ \midrule
CoOp$_{2shots}$      & 74.25\%  & 46.08\%   & 15.10\% & 70.42\% & 53.74\% & 33.33\% & 22.53\% \\
CoOp$_{4shots}$      & 75.19\%  & 46.14\%   & 20.90\% & 72.77\% & 60.44\% & 39.81\% & 20.01\% \\
CoOp$_{8shots}$      & 75.66\%  & 48.41\%   & 28.18\% & 76.10\% & 68.52\% & 46.22\% & 21.02\% \\
CoOp$_{16shots}$     & 75.00\%  & 52.18\%   & 27.14\% & \textbf{78.99}\% & \textbf{75.78}\% & 46.81\% & 25.55\% \\ \bottomrule
\end{tabular}}
\caption{Accuracy comparisons of all comparing methods on CLIP annotated datasets, best performances in bold.}
\label{main}
\end{table}

Our method improves performance over CLIP zero-shot on all experimental datasets and outperforms CoOp on five of the seven datasets, even though CoOp uses human annotated task-relevant labels. On the other two datasets, CoOp performs worse than our method at 2 or 4 shots, and achieves better performances at 8 or 16 shots.

Among the comparing paradigms, only KD and P-WSL can obtain smaller-size deployment model, which is very flexible when facing application scenarios with restrained inference resources. However, student models distilled from the supervision of pre-trained teacher can hardly outperforms the teacher models without task-related training examples.Our method outperforms unsupervised KD on all experimental datasets.

Among P-WSL paradigms, our method outperforms the comparing methods on all datasets except one, on which ALIM-Onehot performs the best. The results demonstrate the effectiveness of our method and partial annotation. CLIP's performance on the SVHN dataset is extremely poor. It predicts almost all samples as "number 0", making it impossible to obtain an effective model using ordinary training algorithms. However, though most prompt templates fail, one of them can achieve an accuracy rate over 25\% and the correctly predicted labels of this prompt template are included into the candidate label sets, which facilitates the training of our method.

\subsection{Synthetic Datasets}

\begin{table}[]
\centering
\setlength{\tabcolsep}{5pt}
\resizebox{\textwidth}{!}{
\begin{tabular}{@{}cccccccccc@{}}
\toprule
\multirow{2}{*}{CIFAR-100} & \multicolumn{3}{c}{$q=0.01$}  & \multicolumn{3}{c}{$q=0.03$}  & \multicolumn{3}{c}{$q=0.05$}  \\ \cmidrule(l){2-10} 
                           & $\eta=0.1$   & $\eta=0.2$   & $\eta=0.3$   & $\eta=0.1$   & $\eta=0.2$   & $\eta=0.3$   & $\eta=0.1$   & $\eta=0.2$   & $\eta=0.3$   \\ \midrule
CC                         & 53.63\% & 48.84\% & 45.50\% & 51.85\% & 47.48\% & 43.37\% & 50.64\% & 45.87\% & 40.87\% \\
RC                         & 52.73\% & 48.59\% & 45.77\% & 52.15\% & 48.25\% & 43.92\% & 46.62\% & 45.46\% & 40.31\% \\
LWC                        & 53.16\% & 48.64\% & 45.51\% & 51.69\% & 47.60\% & 43.39\% & 50.55\% & 45.85\% & 39.83\% \\
LWS                        & 56.05\% & 50.66\% & 45.71\% & 53.59\% & 48.28\% & 42.20\% & 45.46\% & 39.63\% & 33.60\% \\
PiCO                       & 68.27\% & 62.24\% & 58.97\% & 67.38\% & 62.01\% & 58.64\% & 67.52\% & 61.52\% & 58.18\% \\
CR-DPLL                     & 68.12\% & 65.32\% & 62.94\% & 67.53\% & 64.29\% & 61.79\% & 67.17\% & 64.11\% & 61.03\% \\
PiCO+                      & 75.04\% & 74.31\% & 71.79\% & 74.68\% & 73.65\% & 69.97\% & 73.06\% & 71.37\% & 67.56\% \\
IRNet                      & 71.17\% & 70.10\% & 68.77\% & 71.01\% & 70.15\% & 68.18\% & 70.73\% & 69.33\% & 68.09\% \\
ALIM-Scale                 & 77.37\% & 76.81\% & 76.45\% & 77.60\% & 76.63\% & \textbf{75.92}\% & 76.86\% & \textbf{76.44}\% & \textbf{75.67}\% \\
ALIM-Onehot                & 76.52\% & 76.55\% & 76.09\% & 77.27\% & 76.29\% & 75.29\% & \textbf{76.87}\% & 75.23\% & 74.49\% \\
Co-Reg                     & \textbf{78.13}\% & \textbf{78.01}\% & \textbf{77.20}\% & \textbf{77.16}\% & \textbf{76.85}\% & 75.71\% & 76.30\% & 74.91\% & 73.45\% \\ \bottomrule
\end{tabular}}
\caption{Accuracy comparisons on synthetic NPLL datasets, best performances in bold.}
\label{synthetic}
\end{table}

We follow the NPLL dataset generation process used by the previous method \cite{xu2024alim}, which consists of two steps. First, we generate partially labeled datasets by flipping negative labels $\bar{y} \neq y$ to false positive labels with a probability $q = P(\bar{y} \in Y | \bar{y} \neq y)$. Then, we generate noisy partially labeled datasets by randomly substituting the ground-truth label with a non-candidate label with a probability $\eta = P(y \notin Y)$ for each sample. We choose the noise level $\eta$ from \{0.1, 0.2, 0.3\}, and consider $q \in \{0.01, 0.03, 0.05\}$ for CIFAR-100.

We compare our method with six partial label learning (PLL) methods, i.e. CC \cite{DBLP:conf/nips/FengL0X0G0S20}, RC \cite{DBLP:conf/nips/FengL0X0G0S20}, LWC \cite{DBLP:conf/icml/WenCHL0L21}, LWS \cite{DBLP:conf/icml/WenCHL0L21}, PiCO \cite{wang2022pico} and CR-DPLL \cite{wu2022revisiting}, and 4 NPLL methods, i.e. PiCO+ \cite{wang2022pico+}, IRNet \cite{lian2022arnet}, ALIM-Scale and ALIM-Onehot \cite{xu2024alim}. It can be seen that since NPLL methods have the ability to process samples whose real labels are outside their candidate label sets, they generally perform better than PLL methods under noise settings.

On five of the nine subtasks, our method achieves the best performances, while on the remaining subtasks, ALIM-Onehot or ALIM-Scale achieves the best performances (See Table \ref{synthetic}). It is worth noting that our method is not designed for synthetic datasets, but still achieves good performance. It can be clearly seen that our method has more advantages when $q$ is small. This is because there are usually relatively few candidate labels associated with each sample on the dataset annotated by the pre-trained model.

\subsection{Ablations}
\begin{table}[]
\centering
\setlength{\tabcolsep}{6pt}
\resizebox{0.9\textwidth}{!}{
\begin{tabular}{@{}cccccc@{}}
\toprule
Methods   & w/o co-pl & w/o noisy supcont & w/ supcont & w/o prototypical & Co-Reg  \\ \midrule
CIFAR-100 & 68.40\%   & 69.33\%           & 67.96\%    & 69.81\%          & 71.04\% \\ \bottomrule
\end{tabular}}
\caption{Ablation experiments on different degenerations of Co-Reg.}
\end{table}

We conduct experiments on four degenerations of our method to demonstrate the effectiveness of our proposed modules, which are: 1. w/o co-PL: replaces the collaborative pseudo-labeling mechanism to performing pseudo-labeling with their own prediction; 2. w/o noisy supcont: does not perform noisy supervised contrastive learning; 3. w/ supcont: replaces the noisy supervised contrastive learning with traditional supervised contrastive learning; 4. w/o prototypical: does not perform prototypical similarity alignment. It can be seen that, all modules contribute positively to the performance of our method. It is also interesting to observe that, directly employing supervised contrastive learning achieves even worse result due to the noise of pseudo-labels. 

\subsection{Dependencies on Task-Related Samples}
\begin{figure}
    \centering
    \includegraphics[width=0.45\linewidth]{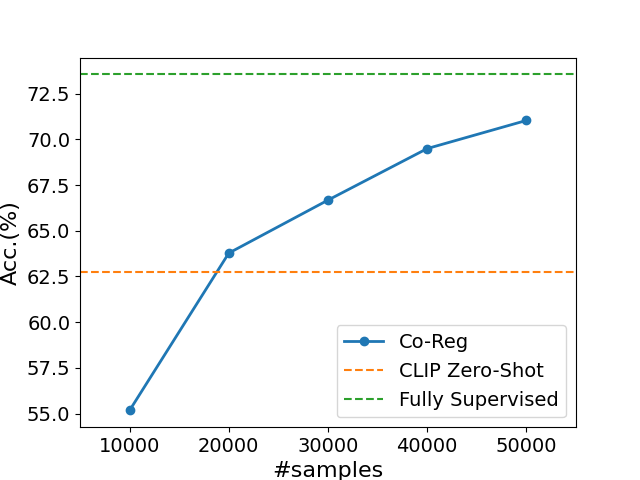}
    \caption{Accuracy on CIFAR-100 using different numbers of unlabeled samples.}
    \label{limits}
\end{figure}

The biggest limitation of our approach in practical applications is the requirement for a large number of downstream unlabeled samples, which we assume are readily available. Here we conduct experiments on CIFAR-100 using different numbers of unlabeled samples and observe the performance of our method to explore its dependence on the number of unlabeled samples. As shown in Fig.\ref{limits}, our method requires 20,000 unlabeled samples (i.e., 200 per class) on CIFAR-100 to exceed the performance of CLIP zero-shot; when using the complete dataset, our method only drops 3\% in performance compared to fully supervised training.

\section{Related Work}
\subsection{Noisy Partial Label Learning}
Partial Label Learning (PLL) \cite{DBLP:conf/ijcai/ZhangY15a, feng2018leveraging, wang2019partial, wang2023deep} is a type of weakly-supervised learning, in which each training sample is associated with multiple candidate labels, and the ground-truth is one of them. The main difficulty of this problem is recognizing the ground-truth label among other associated false-positive candidates, which is called "disambiguation". In recent years, consistency regularization based on data augmentation has been widely applied to PLL, leading to the emergence of a series of impressive methods \cite{wu2022revisiting, wang2022pico, xia2023towards}. They achieved performances with only minor differences compared to fully supervised training, even when the training labels contained a large proportion of false-positive candidates.

Recently, there has been a growing tendency to study a more practical extension of PLL, known as NPLL \cite{lian2022arnet, wang2022pico+, shi2023robust, xu2024alim}, who allowing the existence of noises of partial labels, i.e. the ground-truth is outside of the candidate label set. Since partial labels with noise cannot be fully trusted, researchers typically detect noisy partial labels and perform disambiguation simultaneously during training.

However, previous researches on both PLL and NPLL usually experiment on datasets with randomly flipped false-positive labels, which are relatively easy to disambiguate and exist discrepancy with real-world application scenarios. \cite{xu2021instance} study the problem of instance-dependent PLL whose partial datasets are generated following the prediction of another neural network, simulating the human annotation process. This setup is more comprehensive and challenging, but its practical applicability is still limited.

\subsection{Few-Shot Fine-Tuning and Knowledge Distillation}

To enhance the performance and efficiency of pre-trained vision-language models, several adaptation and distillation techniques have been explored.

One approach is prompt learning \cite{zhou2022learning, zhou2022conditional, khattak2023maple}, which involves fine-tuning the input textual or visual embeddings of vision-language models. This approach can be seen as an extension of manual prompt engineering, which designing appropriate prompts for better aligning downstream images and textual descriptions by domain experts. Since prompt learning only fine-tunes the embedding vectors while keeping the text and image encoder frozen, it requires only few-shot training samples.

Similarly, Adapter \cite{houlsby2019parameter, gao2024clip} has been proposed as a lightweight adaptation mechanism for pre-trained models. Adapters allow the model to learn task-specific information while retaining the majority of its pre-trained parameters. This approach offers a more efficient alternative to fully fine-tuning, enabling quick adaptation to new tasks without the need of large amount of data and significant computational overhead.

Knowledge distillation is a technique where a smaller, student model is trained to replicate the behavior of a larger, teacher model. This approach aims to transfer knowledge from the large, often cumbersome models to more compact and efficient versions, maintaining high performance while reducing computational requirements. \cite{hinton2015distilling} laid the groundwork for this technique, demonstrating that distillation can significantly compress models without substantial loss in accuracy.

\section{Conclusion}
In this paper, we investigate a new paradigm for utilizing pre-trained vision-language model to downstream image classification tasks, i.e. P-WSL, in which the downstream unlabeled samples are annotated automatically by the pre-trained model to train task-specific model and requires no human force. Compared with zero-shot inference, it achieves performance improvements while obtaining much smaller deployment model. We annotate the downstream samples with collections of candidate labels and propose a novel collaborative consistency regularization NPLL method. Experiments show that our method mostly outperforms few-shot fine-tuning technique and other weakly-supervised methods. We hope this article can bring new inspiration to the weakly supervised learning community in the era of large models.

\bibliography{neurips_2024}


\appendix

\section{Additional Experimental Settings}
\subsection{Prompt Templates for CLIP Annotation}\label{prompts}
\begin{verbatim}
prompt_templates_list = [
    "a photo of a {}",
    "a rendering of a {}",
    "a cropped photo of the {}",
    "the photo of a {}",
    "a photo of a clean {}",
    "a photo of a dirty {}",
    "a dark photo of the {}",
    "a photo of my {}",
    "a photo of the cool {}",
    "a close-up photo of a {}",
    "a bright photo of the {}",
    "a cropped photo of a {}",
    "a photo of the {}",
    "a good photo of the {}",
    "a photo of one {}",
    "a close-up photo of the {}",
    "a rendition of the {}",
    "a photo of the clean {}",
    "a rendition of a {}",
    "a photo of a nice {}",
    "a good photo of a {}",
    "a photo of the nice {}",
    "a photo of the small {}",
    "a photo of the weird {}",
    "a photo of the large {}",
    "a photo of a cool {}",
    "a photo of a small {}"
]
\end{verbatim}

\subsection{Class Descriptions for SVHN, FER2013 and GTSRB}\label{names}
\begin{verbatim}
SVHN: ["number 0", "number 1", "number 2", "number 3", "number 4", "number 5", "number 6",
"number 7", "number 8", "number 9"]

FER2013: ["Angry Facial Expression", "Disgust Facial Expression", "Fear Facial Expression",
          "Happy Facial Expression", "Sad Facial Expression", "Surprise Facial Expression",
          "Neutral Facial Expression"]

GTSRB: ["German Traffic Sign: Speed limit 20",
        "German Traffic Sign: Speed limit 30",
        "German Traffic Sign: Speed limit 50",
        "German Traffic Sign: Speed limit 60",
        "German Traffic Sign: Speed limit 70",
        "German Traffic Sign: Speed limit 80",
        "German Traffic Sign: Speed limit 80 cancelled",
        "German Traffic Sign: Speed limit 100",
        "German Traffic Sign: Speed limit 120",
        "German Traffic Sign: No passing",
        "German Traffic Sign: No truck overtaking",
        "German Traffic Sign: Only have priority at the next intersection",
        "German Traffic Sign: Priority road",
        "German Traffic Sign: Give way",
        "German Traffic Sign: Stop",
        "German Traffic Sign: All vehicles prohibited",
        "German Traffic Sign: Trucks prohibited",
        "German Traffic Sign: No entry",
        "German Traffic Sign: Caution!",
        "German Traffic Sign: Curve (to the left)",
        "German Traffic Sign: Curve (to the right)",
        "German Traffic Sign: Continuous curves",
        "German Traffic Sign: Rough road",
        "German Traffic Sign: Slippery road",
        "German Traffic Sign: The road narrows on the right",
        "German Traffic Sign: Construction site",
        "German Traffic Sign: Signal light",
        "German Traffic Sign: Pay attention to pedestrians",
        "German Traffic Sign: Pay attention to children",
        "German Traffic Sign: Watch out for bikes",
        "German Traffic Sign: Watch out for snow/ice on the road",
        "German Traffic Sign: Deer ahead",
        "German Traffic Sign: Unlimited speed",
        "German Traffic Sign: Turn right",
        "German Traffic Sign: Turn left",
        "German Traffic Sign: Go straight",
        "German Traffic Sign: Go straight and turn right",
        "German Traffic Sign: Go straight and turn left",
        "German Traffic Sign: Drive on the right",
        "German Traffic Sign: Drive on the left",
        "German Traffic Sign: Driving around the island",
        "German Traffic Sign: Lift ban on no passing",
        "German Traffic Sign: Lift ban on truck overtaking"]
\end{verbatim}


\end{document}